\documentclass{article}
\usepackage{ijcai19}
\usepackage[english]{babel}

\usepackage{hyperref}
\usepackage{times}
\usepackage{soul}
\usepackage{url}
\usepackage[utf8]{inputenc}
\usepackage[small]{caption}
\usepackage{graphicx}
\usepackage{amsmath}
\usepackage{booktabs}
\usepackage{graphicx}
\usepackage{authblk}
\usepackage{algorithm}
\usepackage[noend]{algpseudocode}
\title{Vulnerability Due to Training Order in Split Learning}

\author[1,2]{\textbf{Harshit Madaan}}
\author[2]{\textbf{Manish Gawali}}
\author[2]{\textbf{Viraj Kulkarni}}
\author[2]{\textbf{Aniruddha Pant}}
\affil[1]{Indian Institute of Science Education and Research, Pune, India}
\affil[2]{DeepTek Inc}

\begin{document}
\maketitle

\begin{abstract}

Split learning (SL) is a privacy-preserving distributed deep learning method used to train a collaborative model without the need for sharing of patient's raw data between clients. In split learning, an additional privacy-preserving algorithm called no-peek algorithm can be incorporated, which is robust to adversarial attacks. The privacy benefits offered by split learning make it suitable for practice in the healthcare domain. However, the split learning algorithm is flawed as the collaborative model is trained sequentially, i.e., one client trains after the other. We point out that the model trained using the split learning algorithm gets biased towards the data of the clients used for training towards the end of a round. This makes SL algorithms highly susceptible to the order in which clients are considered for training. We demonstrate that the model trained using the data of all clients does not perform well on the client's data which was considered earliest in a round for training the model. Moreover, we show that this effect becomes more prominent with the increase in the number of clients. We also demonstrate that the SplitFedv3 algorithm mitigates this problem while still leveraging the privacy benefits provided by split learning.

\end{abstract}

\section{Introduction}

In the healthcare sector, artificial intelligence can help predict a patient’s medical condition using deep learning models. A deep learning model becomes more accurate and robust when it is trained on a large amount of diverse data\cite{hestness1712deep}. It would be ideal if multiple hospitals can cooperate and share their patient’s raw data to train a single model. However, laws concerning data privacy like GDPR, HIPAA, DISHA, etc., impede hospitals from sharing patient’s raw data with any authority. Distributed deep learning methods enable a model to be trained by the collaboration of multiple clients and without compromising the privacy of these clients. For example, In the Federated Learning algorithm, a client only shares the model parameters to a central server instead of the patient’s raw data. Three privacy-preserving distributed deep learning methods are described in the following subsections.

\subsection{Federated Learning}
In Federated learning\cite{konevcny2016federated}\cite{FederatedLearningGoogleAI}\cite{mcmahan2017communication} , the clients train in parallel for a specified number of rounds (or global epochs). During a round, the global model is sent by the server to the clients. Each client that gets the copy of a global model trains it for some local epochs with their respective datasets; thus, each copy of the global model is updated in a different way by different clients. These models are sent back to the server and are then aggregated and averaged on the server. The averaged model is sent back to the clients for validation. This process concludes one round (global epoch) of training.

\subsection{Split Learning}
In split learning\cite{gupta2018distributed}\cite{poirot2019split}, the model architecture is divided into two parts, the front part of the model resides on the client, and the rest of it is on the server. The model is trained sequentially among clients in split learning algorithms. During a round, forward propagation is carried out on the part of the model that resides on the client end, and the last layer's activations are sent to the server where the rest of the model resides. The activations received are used for forward propagation on the server-side model, after which backpropagation is done on the server-side model, and the necessary gradients are sent to the client where backpropagation is carried out on the client-side model. After that, the next client trains the model by communicating with the server, following the same trend of sending activations, receiving gradients, and updating the model. One round (or global epoch) of training completes when all the clients have trained by communicating with the server. Although the server side of the model is the same for every client, there is no notion of the global model in SL as the part of the model architecture that resides on the client-side is unique for each client. If there are `n' clients, then there are `n' different client-side models and a single server-side model.

\subsection{SplitFed}
SplitFed is a hybrid approach between split learning and federated learning. There are two variants of SplitFed proposed by Thapa et al.\cite{thapa2020splitfed}, namely SplitFedv1 and SplitFedv2 and a recent SplitFed approach termed as SplitFedv3 proposed by Gawali et al\cite{gawali2020comparison}. In SplitFed algorithms, the model architecture is divided into segments similar to split learning. The training happens parallelly in SplitFedv1 and SplitFedv3 whereas a model is trained sequentially in SplitFedv2. In SplitFedv2, averaging of the client-side model takes place at the end of each global epoch. In SplitFedv3, the server-side model is averaged out at the end of a global epoch. In SplitFedv1, both parts of the model, i.e., client-side and server-side, are averaged out at the end of a global epoch.

\section{Related work and Vulnerability in SL }

When clients collaborate to train a deep learning model, the model can be trained either sequentially (i.e., one client trains after the other) or parallelly (all the clients train at the same time). When a collaborative model is trained sequentially, it can lead to forgetting when the training set comprises data from non-iid data sources. Each time the model is updated by the data of a client, the model parameters will be updated in such a way to favor the recently seen data. The updated model might forget what was learned when it had trained with the previous client's data resulting in poor classification performance on the `same' dataset. This was shown by Sheller et al.\cite{sheller2018multi} for distributed deep learning methods like IIL (Institutional Incremental Learning) and CIIL (Cyclic Institutional Incremental Learning), where the model is trained sequentially.

In split learning, even though the client-side model is unique for each client, a large trainable portion of the model resides at the server end, and the server-side model is trained sequentially. Forgetting can manifest in split learning such that the performance of the model on the clients that are training early in a cycle could be inadequate.

In the case of federated learning, the global model is an average of all the local models with each client. Each client would have an equal contribution to the final state of the global model when the number of data samples for all clients are the same; hence there is no possibility of forgetting in FL. And even when the number of data samples are uneven, there would still be some contribution from each client.

For the Split-fed approach, SplitFedv2 might also suffer from the problem of forgetting as it also follows the pattern of cyclical training. SplitFedv1 and SplitFedv3, on the other hand, will not show forgetting as the server-side is averaged and does not follow a cyclic pattern. SplitFedv1 has an additional communication overhead where we send all the client-side models to a place where the averaging happens.

\section{Data and Experimental settings}

This section describes the datasets and experimental settings for distributed deep learning methods.

\subsection{Data}

We aggregated the data from three private hospitals (referred to as A, B, C) and two publicly available datasets (MIMIC\cite{johnson2019mimic} and Padchest\cite{bustos2020padchest} referred to as D, E respectively). The data contains chest X-rays with healthy and TB-suspected images of resolution 224x224 pixels. Each chest X-ray is labeled either TB-suspect or TB-negative by a team of expert radiologists. The training set had a total of 8708 X-ray images. In addition to the training set, the validation and testing set contained 2500 images each. The prevalence (percentage of TB-suspect images) of the training set is 50\% and 10\% for the validation set and test set. 
\begin{figure}
\includegraphics[width=\linewidth]{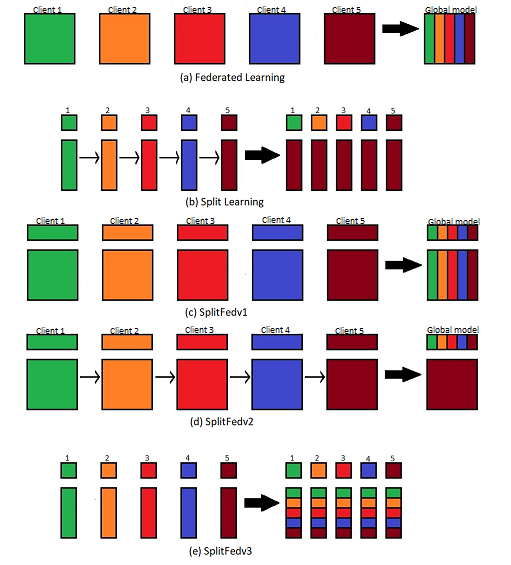}
\caption{Picture the boxes being the state of the model when the model has trained on the respective client dataset. Due to the cyclic pattern of training in split learning, the server-side model's final state is heavily molded according to the requirements of the clients that were considered for training towards the end of the training loop and can result in forgetting. Whereas in a federated setting, the final model is an average of the local models, and thus every client collaborates equally.}
\end{figure}
\begin{table}[hbt!]
\centering
\resizebox{.48\textwidth}{!}{%
\begin{tabular}{ | c | c c c c c c | }
\hline
Data & A & B & C & D & E & Total\\ 
\hline
Train & 1816 & 3772 & 1150 & 880 & 1090 & 8708\\ 
Validation & 500 & 500 & 500 & 500 & 500 & 2500\\ 
Test & 500 & 500 & 500 & 500 & 500 &  2500\\ 
\hline
\end{tabular}
}
\caption{Distribution of chest X-ray images}
\end{table}

\subsection{Experimental settings}

We used Pysyft\cite{ryffel2018generic} library for the experimental implementations. The hospital settings were created using virtual workers, and each virtual worker has access to its own data with no communication to other workers. The Densenet-121\cite{huang2017densely} architecture was used for classification with an Adam optimizer\cite{kingma2014adam} and a learning rate of 1e-4. A combination of threshold dependent (F1-score and Cohen-Kappa\cite{cohen1960coefficient}) and threshold independent (Auprc) metrics are used to evaluate the saved models. We did not consider auc-roc metric due to the prevalence of 10\% in the test set\cite{ozenne2015precision}. The threshold dependent metrics are evaluated on a sensitivity of 0.81. The model with the least validation loss is saved for evaluation. We observed that the model converged within five epochs, and thus, each simulation was executed for ten epochs.

There are many SL configurations\cite{vepakomma2018split}, which can also be extended to SplitFed. The most basic of all is the vanilla configuration. In the vanilla configuration, the clients have to share the labels of the data with the server. A more private configuration called a U-shaped configuration allows the data and the labels to reside on the client. We have used the U-shaped configuration for SL and SplitFed experiments.

\begin{figure}[H]
\includegraphics[width=\linewidth]{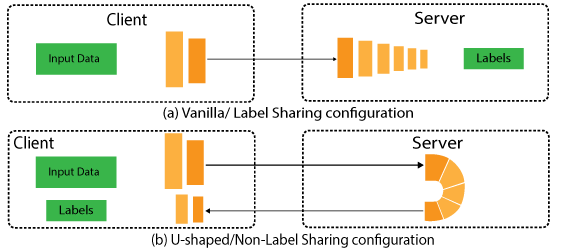}
\caption{In a U-shaped configuration, the input data and labels reside on the client, enhancing privacy over a vanilla configuration where labels have to be shared with the server. }
\end{figure}

\section{Experiments and Results}

The first set of experiments compared the distributed deep learning algorithms, namely Federated Learning, Split Learning, and SplitFedv3, using all the datasets. We have considered two cases for each hospital wherein for the first case, a hospital was the first client, and for the second case, that same hospital trained at the end during a global epoch. For every run, the F1-score, cohen-kappa, and auprc scores were calculated. Table 2, Table 3, and Table 4 show the results for Split Learning, Federated Learning, and SplitFedv3, respectively.

\begin{table}[hbt!]
\centering
\resizebox{.48\textwidth}{!}{%
\begin{tabular}{|l|l|l|l|l|l|l|l|l|l|}
\hline
{Client} & \multicolumn{3}{l|}{AUPRC} & \multicolumn{3}{l|}{F1-score} & \multicolumn{3}{l|}{Kappa} \\ \hline
    & First&  Last& \% drop & First  &   Last   & \% drop&  First  &   Last   &  \% drop       \\ \hline
A&0.4318 &0.5833 &25.97 &0.26 &0.41 &36.59 &0.1107 &0.3032 &63.49  \\ \hline
B&0.4915 &0.8651 &43.19 &0.30 &0.76 &60.53 &0.1607 &0.7340 &78.11  \\ \hline
C&0.5738 &0.7940 &27.73 &0.53 &0.74 &28.38 &0.4528 &0.7064 &35.90  \\ \hline
D&0.4900 &0.6897 &28.95 &0.55 &0.63 &12.70 &0.4794 &0.5788 &17.17  \\ \hline
E&0.7258 &0.8244 &11.96 &0.71 &0.78 &08.97 &0.6787 &0.7510 &09.63  \\ \hline

\end{tabular}
}
\caption{Split Learning results for every client. The second column is the auprc score when that respective client is placed first in training, and the third column is the auprc score when that client is placed at the end of the training loop. The fourth column notes down the percentage change of the performance of the client training early compared to training at the end of a global epoch. A similar pattern is followed for Kappa and F1-scores.}
\end{table}
\begin{table}[hbt!]
\centering
\resizebox{.48\textwidth}{!}{%
\begin{tabular}{|l|l|l|l|l|l|l|l|l|l|}
\hline
{Client} & \multicolumn{3}{l|}{AUPRC} & \multicolumn{3}{l|}{F1-score} & \multicolumn{3}{l|}{Kappa} \\ \hline
    & First&  Last& \% drop &  First  &  Last   & \% drop&  First  &    Last   &  \% drop       \\ \hline
A&0.4482 &0.4482 &- &0.35 &0.35 &- &0.2327 &0.2327 &-\\ \hline
B&0.7646 &0.7646 &- &0.60 &0.60 &- &0.5392 &0.5392 &-  \\ \hline
C&0.7844 &0.7844 &- &0.68 &0.68 &- &0.6428 &0.6428 &-  \\ \hline
D&0.7189 &0.7189 &- &0.66 &0.66 &- &0.6096 &0.6096 &-  \\ \hline
E&0.7656 &0.7656 &- &0.71 &0.71 &- &0.6787 &0.6787 &-  \\ \hline

\end{tabular}
}
\caption{Federated Learning results for every client.}
\end{table}
\begin{table}[hbt!]
\centering
\resizebox{.48\textwidth}{!}{%
\begin{tabular}{|l|l|l|l|l|l|l|l|l|l|}
\hline
{Client} & \multicolumn{3}{l|}{AUPRC} & \multicolumn{3}{l|}{F1-score} & \multicolumn{3}{l|}{Kappa} \\ \hline
    & First& Last& \% drop &  First  &  Last   & \% drop&  First  &    Last   &  \% drop       \\ \hline
A&0.4543 &0.4543 &- &0.34 &0.34 &- &0.2171 &0.2171 &-  \\ \hline
B&0.8063 &0.8963 &- &0.63 &0.63 &- &0.5788 &0.5788 &-  \\ \hline
C&0.6071 &0.6071 &- &0.60 &0.60 &- &0.5454 &0.5454 &-  \\ \hline
D&0.6731 &0.6731 &- &0.67 &0.67 &- &0.6226 &0.6226 &-  \\ \hline
E&0.8060 &0.8060 &- &0.75 &0.75 &- &0.7175 &0.7175 &-  \\ \hline

\end{tabular}
}
\caption{Splitfedv3 results for every client.}
\end{table}
As seen from table 2, In split learning, there was a significant performance drop in all the three performance metrics for each client depending on its order of training. On average, a percentage change of 31\% was observed across all the metrics, with the maximum percentage change being 78\% in the kappa score for client A. Drastic changes were observed for client A as all the three metrics record a percentage change higher than 40\%. Table 3 and Table 4 show that client-order does not matter in federated learning and splitfedv3 as there was no performance drop observed between training early and training late for any client. The clients train the model parallelly in splitfedv3 and federated learning, which makes them immune to any ordering issues.

The next set of experiments tracks the performance of a single client(client A)  over different client settings. For each client setting, the simulations were run for two extreme permutations(one where client A was in the beginning and the other where it trained at the end), and the metric scores of client A were recorded. Initially, only two clients were considered in the training loop, and then the clients were added incrementally one at a time to the training loop. A total of four different client settings were designed for the comparison of performance metrics between the various distributed deep learning algorithms.

\begin{figure}[H]
\includegraphics[width=\linewidth]{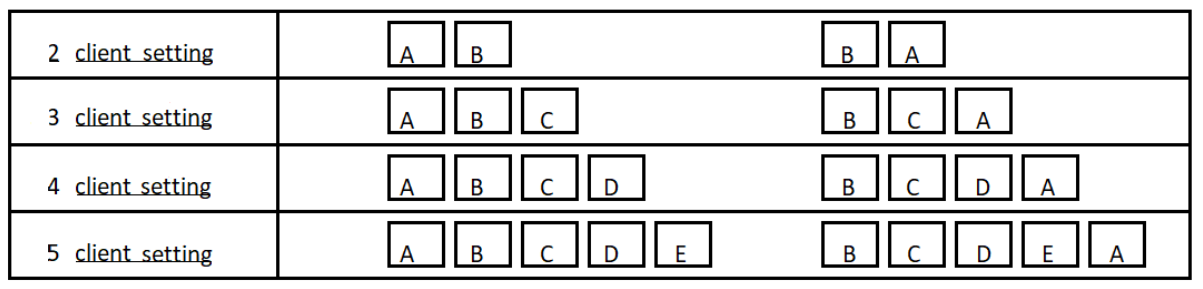}
\caption{Client settings that were considered for the experiments. }
\end{figure}

\begin{table}[hbt!]
\centering
\resizebox{.48\textwidth}{!}{%
\begin{tabular}{|l|l|l|l|l|l|l|l|l|l|}
\hline
{Client A} & \multicolumn{3}{l|}{AUPRC} & \multicolumn{3}{l|}{F1-score} & \multicolumn{3}{l|}{Kappa} \\ \hline
    &  First&  Last& \% drop &  First  & Last   & \% drop&  First  & Last   & \% drop       \\ \hline
2 client setting&0.5986 &0.6616 &09.52 &0.44 &0.51 &13.72 &0.3414 &0.4281 &20.25  \\ \hline
3 client setting&0.5099 &0.6735 &24.29 &0.37 &0.48 &22.92 &0.2604 &0.3905 &33.32  \\ \hline
4 client setting&0.3750 &0.5928 &36.74 &0.31 &0.39 &20.51 &0.1806 &0.2816 &35.87  \\ \hline
5 client setting&0.4318 &0.5833 &25.97 &0.26 &0.41 &36.59 &0.1107 &0.3032 &63.49  \\ \hline

\end{tabular}
}
\caption{Split Learning results for client A over different settings.}
\end{table}
\begin{table}[hbt!]
\centering
\resizebox{.48\textwidth}{!}{%
\begin{tabular}{|l|l|l|l|l|l|l|l|l|l|}
\hline
{Client A} & \multicolumn{3}{l|}{AUPRC} & \multicolumn{3}{l|}{F1-score} & \multicolumn{3}{l|}{Kappa} \\ \hline
    & First& Last&\% drop & First  & Last   & \% drop&  First  &   Last   & \% drop       \\ \hline
2 client setting&0.6198 &0.6198 &- &0.44 &0.44 &- &0.3506 &0.3506 &-  \\ \hline
3 client setting&0.5041 &0.5041 &- &0.42 &0.42 &- &0.3154 &0.3154 &-  \\ \hline
4 client setting&0.3012 &0.3012 &- &0.27 &0.27 &- &0.1272 &0.1272 &-  \\ \hline
5 client setting&0.4482 &0.4482 &- &0.35 &0.35 &- &0.2327 &0.2327 &-  \\ \hline

\end{tabular}
}
\caption{Federated Learning results for client A over different settings}
\end{table}
\begin{table}[hbt!]
\centering
\resizebox{.48\textwidth}{!}{%
\begin{tabular}{|l|l|l|l|l|l|l|l|l|l|}
\hline
{Client A} & \multicolumn{3}{l|}{AUPRC} & \multicolumn{3}{l|}{F1-score} & \multicolumn{3}{l|}{Kappa} \\ \hline
    &  First&  Last& \% drop &  First  & Last   & \% drop&  First  &    Last   &  \% drop       \\ \hline
2 client setting&0.6132 &0.6132 &- &0.45 &0.45 &- &0.3601 &0.3601 &-  \\ \hline
3 client setting&0.5118 &0.5118 &- &0.43 &0.43 &- &0.3354 &0.3354 &-  \\ \hline
4 client setting&0.4734 &0.4734 &- &0.38 &0.38 &- &0.2650 &0.2650 &-  \\ \hline
5 client setting&0.4543 &0.4543 &- &0.34 &0.34 &- &0.2171 &0.2171 &-  \\ \hline

\end{tabular}
}
\caption{SplitFedv3 Learning results for client A over different settings.}
\end{table}

\begin{figure}[H]
\includegraphics[width=\linewidth]{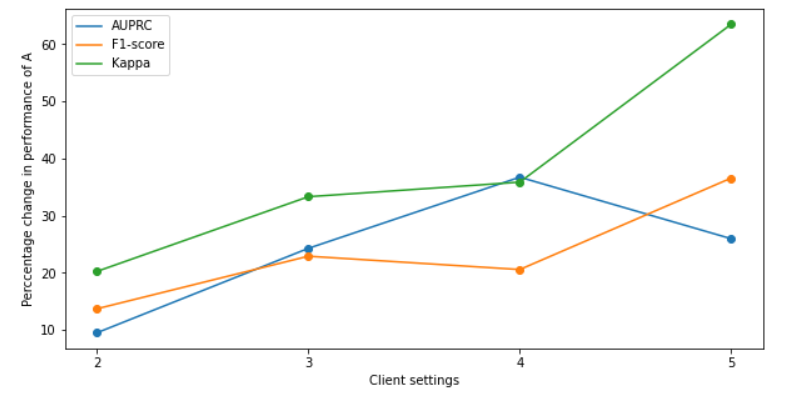}
\caption{Graph of percentage change vs. client settings for split learning results. The x-axis is the number of clients in the particular setting. The y-axis is the percentage change in the performance of client A when its order in the cycle changes from being in the beginning to being at the end. }
\end{figure}
Table 4, Table 5, and Table 6 show the result of the effect of client settings in the distributed deep learning algorithms. As depicted in figure 4, there was an increase in the performance difference of the first client in split learning as more and more clients were added. This shows that forgetting becomes more prominent with an increase in the number of clients. No such performance difference was observed in Federated learning and Splitfedv3. A general decrease in performance was observed for the first client across all the deep learning methods. Although both Splitfedv3 and Federated learning were stable over the ordering changes, the performance results of splitfedv3 were better compared to federated learning for any client setting across all three metrics.

\section{Conclusion}

Split Learning is prone to client-ordering, which is a repercussion of training the collaborative model sequentially. Our results showed that for the split learning algorithm, the performance of the trained model on the data of the client that is placed at the beginning of the training order is compromised while testing. We also showed that the drop in the performance of a client encountered in split learning is absent in other distributed deep learning methods like Federated learning and Splitfedv3 due to parallel training between the clients that are collaborating to train a single model.  Further, we showed that in split learning, the difference between the test performance of a client when training at the earliest and when training at the end of a training loop worsened with an increase in the number of clients in between. SplitFedv3 has a parallel training pattern making it immune to client-ordering issues. It also requires low computation resources on the client-side as only a part of the model is trained on the client-side.  Moreover, split learning algorithms for privacy like no-peek can also be adopted in SplitFedv3.

\bibliographystyle{ieeetr}
\nocite{*}
\bibliography{bibliography}

\begin{thebibliography}{10}

\bibitem{hestness1712deep}
J.~Hestness, S.~Narang, N.~Ardalani, G.~Diamos, H.~Jun, H.~Kianinejad,
  M.~Patwary, Y.~Yang, and Y.~Zhou, ``Deep learning scaling is predictable,
  empirically. arxiv 2017,'' {\em arXiv preprint arXiv:1712.00409}, 2017.

\bibitem{konevcny2016federated}
J.~Kone{\v{c}}n{\`y}, H.~B. McMahan, D.~Ramage, and P.~Richt{\'a}rik,
  ``Federated optimization: Distributed machine learning for on-device
  intelligence,'' {\em arXiv preprint arXiv:1610.02527}, 2016.

\bibitem{FederatedLearningGoogleAI}
B.~McMahan and D.~Rampage, ``Federated learning: Collaborative machine learning
  without centralized training data.''
  \url{https://ai.googleblog.com/2017/04/federated-learning-collaborative.html}.

\bibitem{mcmahan2017communication}
B.~McMahan, E.~Moore, D.~Ramage, S.~Hampson, and B.~A. y~Arcas,
  ``Communication-efficient learning of deep networks from decentralized
  data,'' in {\em Artificial Intelligence and Statistics}, pp.~1273--1282,
  PMLR, 2017.

\bibitem{gupta2018distributed}
O.~Gupta and R.~Raskar, ``Distributed learning of deep neural network over
  multiple agents,'' {\em Journal of Network and Computer Applications},
  vol.~116, pp.~1--8, 2018.

\bibitem{poirot2019split}
M.~G. Poirot, P.~Vepakomma, K.~Chang, J.~Kalpathy-Cramer, R.~Gupta, and
  R.~Raskar, ``Split learning for collaborative deep learning in healthcare,''
  {\em arXiv preprint arXiv:1912.12115}, 2019.

\bibitem{thapa2020splitfed}
C.~Thapa, M.~A.~P. Chamikara, and S.~Camtepe, ``Splitfed: When federated
  learning meets split learning,'' {\em arXiv preprint arXiv:2004.12088}, 2020.

\bibitem{gawali2020comparison}
M.~Gawali, S.~Suryavanshi, H.~Madaan, A.~Gaikwad, B.~P. KN, V.~Kulkarni,
  A.~Pant, {\em et~al.}, ``Comparison of privacy-preserving distributed deep
  learning methods in healthcare,'' {\em arXiv preprint arXiv:2012.12591},
  2020.

\bibitem{sheller2018multi}
M.~J. Sheller, G.~A. Reina, B.~Edwards, J.~Martin, and S.~Bakas,
  ``Multi-institutional deep learning modeling without sharing patient data: A
  feasibility study on brain tumor segmentation,'' in {\em International MICCAI
  Brainlesion Workshop}, pp.~92--104, Springer, 2018.

\bibitem{johnson2019mimic}
A.~E. Johnson, T.~J. Pollard, N.~R. Greenbaum, M.~P. Lungren, C.-y. Deng,
  Y.~Peng, Z.~Lu, R.~G. Mark, S.~J. Berkowitz, and S.~Horng, ``Mimic-cxr-jpg, a
  large publicly available database of labeled chest radiographs,'' {\em arXiv
  preprint arXiv:1901.07042}, 2019.

\bibitem{bustos2020padchest}
A.~Bustos, A.~Pertusa, J.-M. Salinas, and M.~de~la Iglesia-Vay{\'a},
  ``Padchest: A large chest x-ray image dataset with multi-label annotated
  reports,'' {\em Medical image analysis}, vol.~66, p.~101797, 2020.

\bibitem{ryffel2018generic}
T.~Ryffel, A.~Trask, M.~Dahl, B.~Wagner, J.~Mancuso, D.~Rueckert, and
  J.~Passerat-Palmbach, ``A generic framework for privacy preserving deep
  learning,'' {\em arXiv preprint arXiv:1811.04017}, 2018.

\bibitem{huang2017densely}
G.~Huang, Z.~Liu, L.~Van Der~Maaten, and K.~Q. Weinberger, ``Densely connected
  convolutional networks,'' in {\em Proceedings of the IEEE conference on
  computer vision and pattern recognition}, pp.~4700--4708, 2017.

\bibitem{kingma2014adam}
D.~P. Kingma and J.~Ba, ``Adam: A method for stochastic optimization,'' {\em
  arXiv preprint arXiv:1412.6980}, 2014.

\bibitem{cohen1960coefficient}
J.~Cohen, ``A coefficient of agreement for nominal scales,'' {\em Educational
  and psychological measurement}, vol.~20, no.~1, pp.~37--46, 1960.

\bibitem{ozenne2015precision}
B.~Ozenne, F.~Subtil, and D.~Maucort-Boulch, ``The precision--recall curve
  overcame the optimism of the receiver operating characteristic curve in rare
  diseases,'' {\em Journal of clinical epidemiology}, vol.~68, no.~8,
  pp.~855--859, 2015.

\bibitem{vepakomma2018split}
P.~Vepakomma, O.~Gupta, T.~Swedish, and R.~Raskar, ``Split learning for health:
  Distributed deep learning without sharing raw patient data,'' {\em arXiv
  preprint arXiv:1812.00564}, 2018.

\bibitem{singh2019detailed}
A.~Singh, P.~Vepakomma, O.~Gupta, and R.~Raskar, ``Detailed comparison of
  communication efficiency of split learning and federated learning,'' {\em
  arXiv preprint arXiv:1909.09145}, 2019.

\bibitem{abuadbba2020can}
S.~Abuadbba, K.~Kim, M.~Kim, C.~Thapa, S.~A. Camtepe, Y.~Gao, H.~Kim, and
  S.~Nepal, ``Can we use split learning on 1d cnn models for privacy preserving
  training?,'' in {\em Proceedings of the 15th ACM Asia Conference on Computer
  and Communications Security}, pp.~305--318, 2020.

\bibitem{vepakomma2018no}
P.~Vepakomma, T.~Swedish, R.~Raskar, O.~Gupta, and A.~Dubey, ``No peek: A
  survey of private distributed deep learning,'' {\em arXiv preprint
  arXiv:1812.03288}, 2018.

\bibitem{li2019privacy}
W.~Li, F.~Milletar{\`\i}, D.~Xu, N.~Rieke, J.~Hancox, W.~Zhu, M.~Baust,
  Y.~Cheng, S.~Ourselin, M.~J. Cardoso, {\em et~al.}, ``Privacy-preserving
  federated brain tumour segmentation,'' in {\em International Workshop on
  Machine Learning in Medical Imaging}, pp.~133--141, Springer, 2019.

\bibitem{vepakomma2019reducing}
P.~Vepakomma, O.~Gupta, A.~Dubey, and R.~Raskar, ``Reducing leakage in
  distributed deep learning for sensitive health data,'' {\em arXiv preprint
  arXiv:1812.00564}, 2019.

\bibitem{gao2020end}
Y.~Gao, M.~Kim, S.~Abuadbba, Y.~Kim, C.~Thapa, K.~Kim, S.~A. Camtepe, H.~Kim,
  and S.~Nepal, ``End-to-end evaluation of federated learning and split
  learning for internet of things,'' {\em arXiv preprint arXiv:2003.13376},
  2020.

\bibitem{wang2020automated}
P.~Wang, C.~Shen, H.~R. Roth, D.~Yang, D.~Xu, M.~Oda, K.~Misawa, P.-T. Chen,
  K.-L. Liu, W.-C. Liao, {\em et~al.}, ``Automated pancreas segmentation using
  multi-institutional collaborative deep learning,'' in {\em Domain Adaptation
  and Representation Transfer, and Distributed and Collaborative Learning},
  pp.~192--200, Springer, 2020.

\bibitem{sheller2020federated}
M.~J. Sheller, B.~Edwards, G.~A. Reina, J.~Martin, S.~Pati, A.~Kotrotsou,
  M.~Milchenko, W.~Xu, D.~Marcus, R.~R. Colen, {\em et~al.}, ``Federated
  learning in medicine: facilitating multi-institutional collaborations without
  sharing patient data,'' {\em Scientific reports}, vol.~10, no.~1, pp.~1--12,
  2020.

\end{thebibliography}
\end{document}